\documentclass{llncs}

\usepackage{cite}
\usepackage{amsmath,amssymb,amsfonts}
\usepackage{algorithmic}
\usepackage{graphicx}
\usepackage{algorithmic}
\usepackage{algorithm}
\usepackage{multirow}
\usepackage{xcolor}
\usepackage{colortbl}
\usepackage{setspace}
\usepackage{wrapfig}
\usepackage{color}
\usepackage{amsmath,amsfonts,amssymb}
\usepackage{textcomp}
\usepackage{longtable}
\usepackage{booktabs}
\usepackage{ulem}
\def\Reel{\textrm{I\kern-0.21emR}} 
\usepackage{xcolor}
\usepackage{floatflt}

\def\Reel{\textrm{I\kern-0.21emR}} 

\begin{document}

\titlerunning{Predictive K-means with local models}

\title{Predictive K-means with local models}

 \author{Vincent Lemaire\inst{1}, Oumaima Alaoui Ismaili\inst{1}, \\Antoine Cornu\'ejols\inst{2}, Dominique Gay\inst{3}}

 \authorrunning{V.~Lemaire et al.} 

 \institute{Orange Labs, Lannion, France\\
          \and  AgroParisTech, Universit\'e Paris-Saclay,  Paris, France
 \and LIM-EA2525, Universit\'e de La R\'eunion
 }

\maketitle

\begin{abstract}

Supervised classification can be effective for prediction but sometimes weak on interpretability or explainability (XAI). Clustering, on the other hand, tends to isolate categories or profiles that can be meaningful but there is no guarantee that they are useful for labels prediction. Predictive clustering seeks to obtain the best of the two worlds. Starting from labeled data, it looks for clusters that are as pure as possible with regards to the class labels. One technique consists in tweaking a clustering algorithm so that data points sharing the same label tend to aggregate together. With distance-based algorithms, such as k-means, a solution is to modify the distance used by the algorithm so that it incorporates information about the labels of the data points. In this paper, we propose another method which relies on a change of representation guided by class densities and then carries out clustering in this new representation space. We present two new algorithms using this technique and show on a variety of data sets that they are competitive for prediction performance with pure supervised classifiers while offering interpretability of the clusters discovered.

\end{abstract}

\section{Introduction}
\label{intro}

While the power of predictive classifiers can sometimes be awesome on given learning tasks, their actual usability might be severely limited by the lack of interpretability of the hypothesis learned. The opacity of many powerful supervised learning algorithms has indeed become a major issue in recent years. This is why, in addition to good predictive performance as standard goal, many learning methods have been devised to provide readable decision rules \cite{WHI}, degrees of beliefs, or other easy to interpret visualizations. This paper presents a predictive technique which promotes interpretability, explainability as well, in its core design. 

The idea is to combine the predictive power brought by supervised learning with the interpretability that can come from the descriptions of categories, profiles, and discovered using unsupervised clustering. The resulting family of techniques is variously called \textit{supervised clustering} or \textit{predictive clustering.} In the literature, there are two categories of predictive clustering. The first family of algorithms aims at optimizing the trade-off between description and prediction, i.e., aiming at detecting sub-groups in each target class. By contrast, the algorithms in the second category favor the prediction performance over the discovery of all underlying clusters, still using clusters as the basis of the decision function. The hope is that the predictive performance of predictive clustering methods can approximate the performances of supervised classifiers while their descriptive capability remains close to the one of pure clustering algorithms.

Several predictive clustering algorithms have been presented over the years, for instance 
\cite{Eick04,Bilenko,Al-Harbi2006,Cevikalp,cobra2}. 
However, the majority of these algorithms require ($i$) a considerable execution time, and ($ii$) that numerous user parameters be set. In addition, some algorithms are very sensitive to the presence of noisy data and consequently their outputs are not easily interpretable (see \cite{bookpct} for a survey).
This paper presents a new predictive clustering algorithm. The underlying idea is to use any existing distance-based clustering algorithms, e.g. k-means, but on a redescription space where the target class is integrated. 
The resulting algorithm has several desirable properties: there are few parameters to set, its computational complexity is almost linear in $m$, the number of instances, it is robust to noise, its predictive performance is comparable to the one obtained with classical supervised classification techniques and it tends to  produce groups of data that are easy to interpret for the experts. 

The remainder of this paper is organized as follows: Section II introduces the basis of the new algorithm, the computation of the clusters, the initialization step and the classification that is realized within each cluster. The main computation steps of the resulting predictive clustering algorithms are described in Algorithm \ref{algo}. We then report experiments that deal with the predictive performance in Sections \ref{sec:experiments}. We focus on the supervised classification performance  to assess if predictive clustering could reach the performances of algorithms dedicated to supervised classification. Our algorithm is compared using a variety of data sets with powerful supervised classification algorithms in order to assess its value as a predictive technique. And an analysis of the results is carried out. Conclusion and perspectives are discussed in Section \ref{conclusion}. 


\vspace{-2mm}
\section{Turning the K-means algorithm predictive}
\label{algo:principle}
\vspace{-2mm}

The k-means algorithm is one of the simplest yet most commonly used clustering algorithms. It seeks to partition $m$ instances ($X_1,\ldots X_m$) into $K$ groups ($B_1, \ldots, B_K$) so that instances which are close are assigned to the same cluster while clusters are as dissimilar as possible. The objective function can be defined as:
{\fontsize{8}{10}\selectfont
\begin{equation}\label{eq_kmeans}
{\cal G} \; = \, \operatornamewithlimits{Argmin}_{B_i} \sum_{i=1}^K \sum_{X_j \in B_i} \Vert X_j - \mu_i \Vert^2 
\end{equation}}
where $\mu_i$ are the centers of clusters $B_i$ and we consider the Euclidean distance.

Predictive clustering adds the constrain of maximizing clusters purity (i.e. instances in a cluster should share the same label). In addition, the goal is to provide results that are easy to interpret by the end users. 
The objective function of Equation (\ref{eq_kmeans}) needs to be modified accordingly. 

One approach is to modify the distance used in conventional clustering algorithm in order to incorporate information about the class of the instances. This modified distance should make points differently labelled appear as more distant than in the original input space. Rather than modifying the distance, one can instead alter the input space. This is the approach taken in this paper, where the input space is partitioned according to class probabilities prior to the clustering step, thus favoring clusters of high purity.
Besides the introduction of a technique for computing a new feature space, we propose as well an adapted initialization method for the modified k-means algorithm. We also show the advantage of using a specific classification method within each discovered cluster in order to improve the classification performance. The main steps of the resulting algorithm are described in Algorithm \ref{algo}. In the remaining of this section II we show how each step of the usual K-means is modified to yield a predictive clustering algorithm.
\vspace{-4mm}
\begin{algorithm}
\begin{algorithmic}
\fontsize{9}{11}\selectfont
\STATE \textbf{Input:}\\
\STATE - $D$: a data set which contains $m$ instances. Each one ($X_i$) $i \in \lbrace 1,\ldots,m\rbrace$ is described by $d$ descriptive features and a label $C_i  \in \lbrace 1,\ldots, J \rbrace.$
\STATE - $K$: number of clusters .
\STATE \textbf{Start:}\\
\STATE 1) Supervised preprocessing of data to represent each $X_i$ as $\widehat{X}_i$ in a new feature space $\Phi({\cal X})$.
\STATE 2) Supervised initialization of centers.
\REPEAT
\STATE 3) {\it Assignment:} generate a new partition by assigning each instance $ \widehat{X}_i $ to the nearest cluster.
\STATE 4) Representation: calculate the centers of the new partition.
\UNTIL{the convergence of the algorithm}
\STATE 5) Assignment classes to the obtained clusters: 
\STATE  \hspace{1.3cm}       - {\it method 1}: majority vote.
\STATE  \hspace{1.3cm}       - {\it method 2}: local models.
\STATE 6) Prediction the class of the new instances in the deployment phase:
\STATE \hspace{1.3cm} $\rightarrow$ \textit{the closest cluster class (if {\it method 1} is used).}
\STATE \hspace{1.3cm} $\rightarrow$ \textit{local models (if {\it method 2} is used).}
\STATE \textbf{End}
\end{algorithmic}
\caption{Predictive K-means algorithm}\label{algo}
\end{algorithm}
\vspace{-4mm}

\begin{figure}[htbp!]
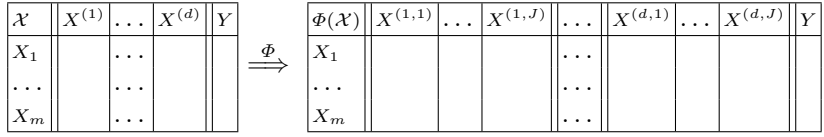

    \centering
    {\fontsize{7}{9}
    \begin{tabular}{|l||c|c|c||c|}
    \hline
    $\cal X$ &  $X^{(1)}$ & \ldots & $X^{(d)}$ & $Y$\\
    \hline
    $X_1$ &   & \ldots &  & \\
    \ldots &   & \ldots &  & \\
    $X_m$ &   & \ldots & & \\
    \hline
    \end{tabular}}
    $\overset{\Phi}{\Longrightarrow}$
    {\fontsize{7}{9}
    \begin{tabular}{|l||c|c|c||c||c|c|c||c|}
    \hline
    $\Phi({\cal X})$ &  $X^{(1,1)}$ &  \ldots &  $X^{(1,J)}$ & \ldots &  $X^{(d,1)}$ &  \ldots &  $X^{(d,J)}$ & $Y$\\
    \hline
    $X_1$ & & &  & \ldots & & & & \\
    \ldots & & &  & \ldots & & & & \\
    $X_m$ & & &  & \ldots & & & & \\
    \hline
    \end{tabular}}
    \caption{$\Phi$ redescription scheme from $d$ variables to $d\times J$ variables, with log-likelihood values: $\log P(X^{(n)} | C_j)$}
    \label{fig:process}
    \vspace{-4mm}
\end{figure}

\vspace{-2mm}
\noindent \textbf{A modified input space for predictive clustering - }The principle of the proposed approach is to partition the input space according to the class probabilities $P(C_j|X)$. More precisely, let the input space ${\cal X}$ be of dimension $d$, with numerical descriptors as well as categorical ones. An example $X_i \in {\cal X}$ ($X_i = [X_i^{(1)}, \ldots, X_i^{(d)}]^\top$) will be described in the new feature space $\Phi({\cal X})$ by $d \times J$ components, with $J$ being the number of classes. Each component $X_i^{(n)}$ of $X_i \in {\cal X}$ will give $J$ components $X_i^{(n,j)}$, for $j\in \{1, \ldots, J\}$,  of the new description $\widehat{X}_i$ in $\Phi({\cal X})$, where $\widehat{X}_i^{(n,j)}  =  \log P(X^{(n)} = X_i^{(n)} | C_j)$, i.e., the log-likelihood values. Therefore, an example $X$ is redescribed according to the (log)-probabilities of observing the values of original input variables given each of the $J$ possible classes (see Figure~\ref{fig:process}). Below, we describe a method for computing these values. But first, we analyze one property of this redescription in $\Phi({\cal X})$ and the distance this can provide. 

\vspace{-2mm}

\noindent \textbf{Property of the modified distance - } Let us denote $dist_B^p$ the new distance defined over $\Phi({\cal X})$. For the two recoded instances $\hat{X_1}$ and $\hat{X_2} \in \Reel^{d \times J}$, the formula of $dist_B^p$ is (in the following we omit $\widehat{X} = \widehat{X}_i$ in the probability terms for notation simplification):
{\fontsize{9}{11}\selectfont
\begin{equation}
 dist_B^p(\hat{X_1},\hat{X_2})= \sum_{j=1}^J\parallel \log(P(\hat{X_1}|C_j)) - \log(P(\hat{X_2}|C_j))\parallel_p
\end{equation}}
where $\parallel.\parallel_p$ is a Minkowski distance.  Let us denote now, $\Delta^p(\hat{X_1},\hat{X_2})$ the distance between the (log)-posterior probabilities of two instances $\hat{X_1}$ and $\hat{X_2}$.
The formula of this distance as follow:
{\fontsize{9}{11}\selectfont
\begin{equation}
\label{dist_distribution}
\Delta^p(\hat{X_1},\hat{X_2}) = \sum_{j=1}^J\parallel \log(P(C_j|\hat{X_1})) - \log(P(C_j|\hat{X_2}))\parallel_p     
\end{equation}
where $\forall i \in \lbrace 1,\ldots,m\rbrace$, $P(C_j|\hat{X_i}) = \frac{P(C_j)\prod_{n=1}^d P(X_{i}^{(n)}|C_j)}{P(\hat{X_i})}$ }
(using the hypothesis of features independence conditionally to the target class). From the distance given in equation \ref{dist_distribution}, we find the following inequality:

{\fontsize{8}{10}\selectfont

\begin{equation}
\label{inequality}
\Delta^p(\hat{X_1},\hat{X_2}) \leq \bigl\{dist_B^p(\hat{X_1},\hat{X_2}) + J\parallel \log(P(\hat{X_2}))- \log(P(\hat{X_1})) \parallel \bigr\}
\end{equation}

\begin{proof}
\begin{align*}
\Delta^p &= \sum_{j=1}^J\parallel \log(P(C_j|\hat{X_1})) - \log(P(C_j|\hat{X_2}))\parallel_p \\
        &= \sum_{j=1}^J\parallel \log(\frac{P(\hat{X_1}|C_j)P(C_j)}{P(\hat{X_1})}) - \log(\frac{P(\hat{X_2}|C_j)P(C_j)}{P(\hat{X_2})})\parallel_p \\
        &= \sum_{j=1}^J\parallel \log(P(\hat{X_1}|C_j)) - \log(P(\hat{X_1}))   - \log(P(\hat{X_2}|C_j)) + \log(P(\hat{X_2}))  \parallel_p\\
        &\leq  \sum_{j=1}^J[ A + B] 
\end{align*}

\noindent with  $\Delta^p = \Delta^p(\hat{X_1},\hat{X_2}) $
and  $A = \parallel \log(P(\hat{X_1}|C_j)) - \log(P(\hat{X_2}|C_j))\parallel_p$\\
and ~$B = \parallel \log(P(\hat{X_2})) - \log(P(\hat{X_1})) \parallel_p$\\
then $\Delta^p \leq dist_B^p(\hat{X_1},\hat{X_2}) + J\parallel \log(P(\hat{X_2})) - \log(P(\hat{X_1}))\parallel_p$.
\end{proof}}

This above inequality expresses that two instances that are close in terms of distance $dist_B^p$ will also be close in terms of their probabilities of belonging to the same class. Note that the distance presented above can be integrated into any distance-based clustering algorithms.\\

\vspace{-2mm}
\noindent \textbf{Building the log-likelihood redescription $\Phi({\cal X})$  - }Many methods can estimate the new descriptors $X_i^{(n,j)} = \log P(X^{(n)} = X_i^{(n)} | C_j)$ from a set of examples. In our work, we use a supervised discretization method for numerical attributes and a supervised grouping values for categorical attributes to obtain respectively intervals and group values in which $P(X^{(n)} = X_i^{(n)} | C_j)$ could be measured. 
 The used supervised discretization method is described in \cite{BoulleML06} and the grouping method in \cite{BoulleJMLR05}. The two methods have been compared with extensive experiments to corresponding state of the art algorithms.
These methods computes univariate partitions of the input space using supervised information. It determines the partition of the input space to optimize the prediction of the labels of the examples given the intervals in which they fall using the computed partition. The method finds the best partition (number of intervals and thresholds) using a Bayes estimate. An additional bonus of the method is that outliers are automatically eliminated and missing values can be imputed. \\

\vspace{-2mm}
\noindent \textbf{Initialisation of centers - }Because clustering is a NP-hard problem, heuristics are needed to solve it, and the search procedure is often iterative, starting from an initialized set of prototypes. One foremost example of many such distance-based methods is the k-means algorithm. It is known that the initialization step can have a significant impact both on the number of iterations and, more importantly, on the results which correspond to local minima of the optimization criterion (such as Equation~\ref{eq_kmeans} in \cite{Meila1998}). 
However, by contrast to the classical clustering methods, predictive clustering can use supervised information for the choice of the initial prototypes. In this study, we chose to use the \texttt{K++R} method. Described in \cite{LemaireIJCNN2015initialisation}, it follows an ``exploit and explore'' strategy where the class labels are first exploited before the input distribution is used for exploration in order to get the apparent best initial centers. The main idea of this method is to dedicate one center per class (comparable to a \textit{``Rocchio"} \cite{Manning2008} solution). Each center is defined as the average vector of instances which have the same class label. If the predefined number of clusters ($K$) exceeds the number of classes ($J$), the initialization continues using the K-means++ algorithm~\cite{Arthur2007} for the $K-J$ remaining centers in such a way to add diversity. This method can only be used when  $K \geq J$, but this is fine since in the context of supervised clustering\footnote{~In the context of supervised clustering, it does not make sense to cluster instances in $K$ clusters where $K  <  J$} we do not look for clusters where $K  <  J$. The complexity of this scheme is ${\cal O}(m+(K-J)m) < {\cal O}(mK)$, where $m$ is the number of examples. When $K=J$, this method is deterministic.\\

\vspace{-2mm}
\noindent \textbf{Instance assignment and centers update - }Considering the Euclidean distance and the original K-means procedure for updating centers, at each iteration, each instance is assigned to the nearest cluster ($j$) using the $\ell_2$ metric ($p=2$) in the redescription space $\Phi({\cal X})$. The $K$ centers are then updated according to the K-Means procedure. This choice of distance (Euclidean) in the adopted k-means strategy could have an influence on the (predictive) relevance of the clusters but has not been studied in this paper. \\

\vspace{-2mm}
\noindent \textbf{Label prediction in predictive clustering - }Unlike classical clustering which aims only at providing a description of the available data, predictive clustering can also be used in order to make prediction about new incoming examples that are unlabeled. 

The commonest method used for prediction in predictive K-means is the majority vote. A new example is first assigned to the cluster of its nearest prototype, and the predicted label is the one shared by the majority of the examples of this cluster. This method is not optimal. Let us call $P_M$ the frequency of the majority class in a given cluster. The true probability $\mu$ of this class obeys the Hoeffding inequality: $P\bigl(|P_M - \mu | \geq~\varepsilon \bigr) \; \leq \; 2 \exp(- 2 \, m_k \; \varepsilon^2)$  with $m_k$ the number of instances assigned to the cluster $k$. If there are only 2 classes, the error rate is $1 - \mu$ if $P_M$ and $\mu$ both are $> 0.5$. But the error rate can even exceed 0.5 if $P_M > 0.5$ while actually $\mu < 0.5$. The analysis is more complex in case of more than two classes. It is not the object of this paper to investigate this further. But it is apparent that the majority rule can often be improved upon, as is the case in classical supervised learning.

Another evidence of the limits of the majority rule is provided by the examination of the ROC curve~\cite{flach2012machine}. Using the majority vote to assign classes for the discovered clusters generates a ROC curve where instances are ranked depending on the clusters. Consequently, the ROC curve presents a sequence of steps.  The area under the ROC curve is therefore suboptimal compared to a ROC curve that is obtained from a more refined ranking of the examples, e.g., when class probabilities are dependent upon each example, rather than groups of examples. 

One way to overcome these limits is to use local prediction models in each cluster, hoping to get better prediction rules than the majority one. However, it is necessary that these local models: 
1) can be trained with few instances, 
2) do not overfit,
3) ideally, would not imply any user parameters to avoid the need for local cross-validation,
4) have a linear algorithmic complexity $O(m)$ in a learning phase, where $m$ is the number of examples,
5) are not used in the case where the information is insufficient and the majority rule is the best model we can hope for,
6) keep (or even improve) the initial interpretation qualities of the global model.
Regarding item (1), a large study has been conducted in~\cite{SalperwyckIJCNN2011learning} in order to test the prediction performances in function of the number of training instance of the most commonly classifiers. One prominent finding was that the Naive Bayes (NB) classifier often reaches good prediction performances using only few examples (Bouchard \& Triggs's study~\cite{Bouchard2004} confirms this result). This fact remains valid even when features receive weights ({\it e.g.,} Averaging Naive Bayes (ANB) and  Selective Naive Bayes (SNB)~\cite{Langley1994}). We defer discussion of the other items to the Section \ref{sec:experiments} on experimental results.  

\smallskip
In our experiments, we used the following procedure to label each incoming data point~$X$:
i) $X$ is redescribed in the space $\Phi({\cal X})$ using the method described in Section II, 
ii) $X$ is assigned to the cluster $k$ corresponding to the nearest center
iii) the local model, $l$, in the corresponding cluster is used to predict the class of $X$ (and the probability memberships) if a local model exits:
$ P(j|X) = \text{argmax}_{1\leq j\leq J}(P_{SNB_l}(C_j|X))$ otherwise the majority vote is used (Note: $P_{SNB_l}(C_j|X))$ is described in the next section).


\section{Comparison with supervised Algorithm}
\label{sec:experiments}
\vspace{-2mm}

\subsection{The chosen set of classifiers}
\label{usedclassifiers}

To test the ability of our algorithm to exhibit high predictive performance while at the same time being able to uncover interesting clusters in the different data sets, we have compared it with three powerful classifiers (in the spirit of, or close to our algorithm) from the state of the art: Logistic Model Tree (\texttt{LMT}) \cite{LMT}, Naives Bayes Tree (\texttt{NBT}) \cite{Kohavi96scalingup} and Selective Naive Bayes (SNB) \cite{BoulleJMLR07}. This section briefly described these classifiers.

{\bf $\bullet$ Logistic Model Tree (\texttt{LMT})} \cite{LMT} combines logistic regression and decision trees. It seeks to improve the performance of decision trees. Instead of associating each leaf of the tree to a single label and a single probability vector (piecewise constant model), a logistic regression model is trained on the instances assigned to each leaf to estimate an appropriate vector of probabilities for each test instance (piecewise linear regression model).  The logit-Boost algorithm is used to fit a logistic regression model at each node, and then it is partitioned using information gain as a function of impurity.

{\bf $\bullet$ Naives Bayes Tree (\texttt{NBT})} \cite{Kohavi96scalingup} is a hybrid algorithm, which deploys a naive Bayes classifier on each leaf of the built decision tree. \texttt{NBT} is a classifier which has often exhibited good performance compared to the standard decision trees and naive Bayes classifier. 

{\bf $\bullet$ Selective Naive Bayes (\texttt{SNB})} is a variant of \texttt{NB}. 
One way to average a large number of selective naive Bayes classifiers obtained with different subsets of features is to use one model only, but with features weighting \cite{BoulleJMLR07}. The Bayes formula under the hypothesis of features independence conditionally to classes becomes:   $P(j|X)=\frac{P(j)\prod_{f}P(X^{f}|j)^{W_f}}{\sum_{j=1}^{K}\left[P(j)\prod_{f}P(X^{f}|j)^{W_f}\right]}$, where $W_f$ represents the weight of the feature $f$, $X^f$ is component $f$ of $X$, $j$ is the class labels. The predicted class $j$ is the one that maximizes the conditional probability $P(j|X)$. The probabilities $P(X_i|j)$ can be estimated by interval using a discretization for continuous features. For categorical features, this estimation can be done if the feature has few different modalities.  Otherwise,  grouping into modalities is used.
The resulting algorithm proves to be quite efficient on many real data sets~\cite{Hand2001}. 

{\bf  $\bullet$ Predictive K-Means ($\texttt{PKM}_{\texttt{MV}}$, $\texttt{PKM}_{\texttt{SNB}}$):} (i) $\texttt{PKM}_{\texttt{VM}}$ corresponds to the Predictive K-Means described in Algorithm \ref{algo} where prediction is done according to the  Majority Vote; (ii) $\texttt{PKM}_{\texttt{SNB}}$ corresponds to the Predictive K-Means described in Algorithm \ref{algo} where prediction is done according to a local classification model. 

{\bf  $\bullet$ Unsupervised K-Means ($ \texttt{KM}_{\texttt{MV}}$)} is the usual unsupervised K-Means with prediction done using the Majority Vote in each cluster. This classifier is given for comparison  as a baseline method. The pre-processing is not supervised and the initialization used is k-means++~\cite{Arthur2007} (in this case since the initialization is not deterministic we run k-means 25 times and we keep the best initialization according to the Mean Squared Error). Among the existing unsupervised pre-processing approaches~\cite{Milligan1988}, depending on the nature of the features, continuous or categorical, we used:

\begin{itemize}
\item for Numerical attribute: Rank Normalization (RN). The purpose of rank normalization is to rank continuous feature values and then scale the feature into $[0,1]$. The different steps of this approach are: ($i$) rank feature values $u$ from lowest to highest values and then divide the resulting vector into $H$ intervals, where $H$ is the number of intervals, ($ii$) assign for each interval a label $r \in \{1,...,H\}$ in increasing order, ($iii$) if $X_{iu}$ belongs to the interval $r$, then  $X'_{iu} = \frac{r}{H}$. In our experiments, we use $H=100$.

\smallskip
\item for Categorical attribute: we chose to use a Basic Grouping Approach (BGB). It aims at transforming feature values into a vector of Boolean values. The different steps of this approach are: ($i$) group feature values into $g$ groups with as equal frequencies as possible, where $g$ is a parameter given by the user, ($ii$) assign for each group a label $r \in \{1,...,g\}$, ($iii$) use a full disjunctive coding. In our experiments, we use $g=10$.
\end{itemize}

\begin{figure}
\centering
\includegraphics[width=0.65\linewidth]{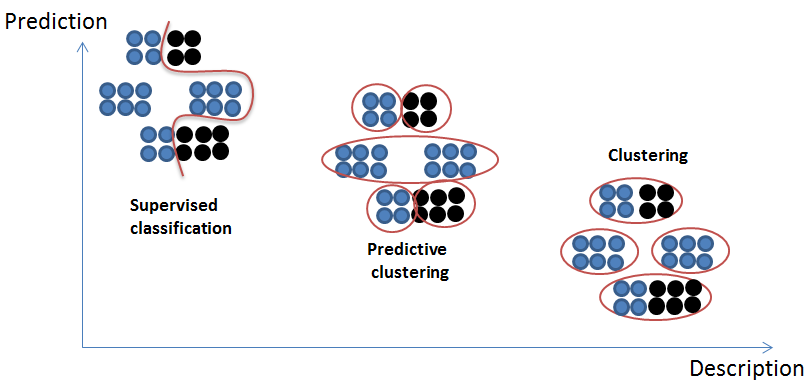}
\caption{Differences between the three types of ``classification'' \label{pig_clustering_predictif}}
\end{figure}

\vspace{-4mm}
In the Figure \ref{pig_clustering_predictif}
we suggest a two axis figure to situate the algorithms described above: a vertical axis for their ability to describe (explain) the data (from low to high) and horizontal axis for their ability to predict the labels (from low to high). In this case the selected classifiers exemplify various trade-offs between prediction performance and explanatory power: ($i$) $\texttt{KM}_{\texttt{MV}}$ more dedicated to description would appear in the bottom right corner; ($ii$) \texttt{LMT}, \texttt{NBT} and \texttt{SNB} dedicated to prediction would go on the top left corner; and ($iii$) $\texttt{PKM}_{\texttt{VM}}$, $\texttt{PKM}_{\texttt{SNB}}$  would lie in between. Ideally, our algorithm,  $\texttt{PKM}_{\texttt{SNB}}$ should place itself on the top right quadrant of this kind of figure with both good prediction and description performance.

Note that in the reported experiments, $K = J$ ({\it i.e,} number of clusters = number of classes). This choice which biases the algorithm to find one cluster per class, is detrimental for predictive clustering, thus setting a lower bound on the performance that can be expected of such an approach.

\subsection{Experimental protocol}
The comparison of the algorithms have been performed on 8 different datasets of the UCI repository \cite{Lichman2013}. These datasets were chosen for their diversity in terms of classes, features (categorical and numerical) and instances number (see Table~\ref{used_data}).  

\vspace{-0.55cm}
\begin{table}[h!]
\fontsize{8}{10}\selectfont
\begin{center}
\begin{tabular}{|c||c|c|c|c||c|c||c|c|c|c|}
\hline  {\bf Datasets} & Instances & ${\# } V_n$ & ${\#}V_c$ & ${\#}$ Classes & {\bf Datasets} & Instances & ${\# } V_n$ & ${\#}V_c$ & ${\#}$ Classes \\
\hline Glass & 214 & 10 &0 & 6 & Waveform & 5000 & 40  & 0  & 3 \\
\hline Pima & 768 & 8 & 0 & 2 & Mushroom & 8416 & 0 & 22 & 2 \\
\hline Vehicle & 846 & 18 & 0 & 4 & Pendigits & 10992  & 16  & 0 & 10 \\
\hline Segmentation & 2310 &  19  & 0 &  7 & Adult & 48842 & 7 & 8 & 2\\\hline
\end{tabular}
\medskip
\caption{The used datasets, $V_n$: numerical features, $V_c$: categorical features.} \label{used_data}
\end{center}
\vspace{-17mm}
\end{table}

\subsubsection{Evaluation of the performance:} In order to compare the performance of the algorithms presented above, the same folds in the train/test have been used. The results presented in Section \ref{results} are those obtained in the test phase using a $10\times 10$ folds cross validation (stratified). The predictive performance of the algorithms are evaluated using the AUC (area under the ROC's curve). It is computed as follows:   $\texttt{AUC} = \sum_{i}^{C} P(C_i) \texttt{AUC}(C_i)$, where $\texttt{AUC}(i)$ denotes the AUC's value in the class $i$ against all the others classes and $P(Ci)$ denotes the prior on the class $i$ (the elements frequency in the class $i$).  $\texttt{AUC} (i)$ is calculated using the probability vector $P(C_i|X)$  $\forall i$.


\vspace{-2mm}
\begin{table}[h!]
\fontsize{7}{9}\selectfont
\begin{center}
\begin{tabular}{|c||c|c|c|c|c|c|}\hline
\multicolumn{7}{|c|}{average results (in the test phase) using \textbf{ACC}} \\
\hline  {\bf Data} & $\texttt{KM}_{\texttt{MV}}$ & $\texttt{PKM}_{\texttt{MV}}$ & $\texttt{PKM}_{\texttt{SNB}}$  & $\texttt{LMT}$ & \texttt{NBT} & \texttt{SNB}\\
\hline Glass 		& $70.34 \pm 8.00$ & $89.32 \pm 6.09$ & \cellcolor{orange!15}$95.38 \pm 4.66$ & $97.48 \pm 2.68$ & $94.63 \pm 4.39$ & \textbf{97.75} $\pm 3.33$\\  
\hline Pima			& $65.11 \pm 4.17$ & $66.90 \pm 4.87$ & \cellcolor{orange!15}$73.72 \pm 4.37$ & \textbf{76.85} $\pm 4.70$ & $75.38 \pm 4.71$ & $75.41 \pm 4.75$\\ 
\hline Vehicle		& $37.60 \pm 4.10$ & $47.35 \pm 5.62$ & \cellcolor{orange!15}$72.21 \pm 4.13$ & $\textbf{82.52} \pm 3.64$ & $70.46 \pm 5.17$ & $64.26 \pm 4.39$ \\ 
\hline Segment		& $67.50 \pm 2.35$ & $80.94 \pm 1.93$ & \cellcolor{orange!15}$96.18 \pm 1.26$ & $\textbf{96.30} \pm 1.15$ & $95.17 \pm 1.29$ & $94.44 \pm 1.48$ \\ 
\hline Waveform		& $50.05 \pm 1.05$ & $49.72 \pm 3.39$ & \cellcolor{orange!15}$84.04 \pm 1.63$ & $\textbf{86.94} \pm 1.69$ & $79.87 \pm 2.32$  & $83.14 \pm 1.49 $\\ 
\hline Mushroom		& $89.26 \pm 0.97$ & $98.57 \pm 3.60$ & \cellcolor{orange!15}$\textbf{99.94} \pm 0.09$ & $98.06 \pm 4.13$ & $95.69 \pm 6.73$ & $99.38 \pm 0.27$ \\ 
\hline PenDigits	& $73.65 \pm 2.09$ & $76.82 \pm 1.33$ & \cellcolor{orange!15}$97.35 \pm 1.36$	& $\textbf{98.50} \pm 0.35$ &  $95.29 \pm 0.76$ & $89.92 \pm 1.33$ \\ 
\hline Adult		& $76.07 \pm 0.14$ & $77.96 \pm 0.41$ & \cellcolor{orange!15}$\textbf{86.81} \pm 0.39$ & $83.22 \pm 1.80$ & $79.41 \pm 7.34$ & $86.63 \pm 0.40$\\\hline
\hline Average		& $66.19$ & $73.44$ & \cellcolor{orange!25}$88.20$ & $\textbf{89.98}$ & $85.73$ & $86.36$ \\\hline
\multicolumn{7}{|c|}{average results (in the test phase) using 100 x \textbf{AUC}} \\

\hline  {\bf Data} & $\texttt{KM}_{\texttt{MV}}$ & {\bf $\texttt{PKM}_{\texttt{MV}}$} & {\bf $\texttt{PKM}_{\texttt{SNB}}$ } &  $\texttt{LMT}$  & $\texttt{NBT}$ &  \texttt{SNB}\\
\hline Glass 		& $85.72 \pm 5.69$ & $96.93 \pm 2.84$ & \cellcolor{orange!15}$98.27 \pm 2.50$ & $97.94 \pm 0.19$ & $98.67 \pm 2.05$ & $\textbf{99.77} \pm 0.54$\\ 
\hline Pima 		& $65.36 \pm 5.21$ & $65.81 \pm 6.37$ & \cellcolor{orange!15}$78.44 \pm 5.35$ & $\textbf{83.05} \pm 4.61$ & $80.33 \pm 5.21$ &$80.59 \pm 4.78$\\ 
\hline Vehicle 		& $65.80 \pm 3.36$ & $74.77 \pm 3.14$ & \cellcolor{orange!15}$91.15 \pm 1.75$ & $\textbf{95.77} \pm 1.44$ & $88.07 \pm 3.04$ & $87.19 \pm 1.97$\\ 
\hline Segment 		& $91.96 \pm 0.75$ & $95.24 \pm 0.75$ & \cellcolor{orange!15}$99.51 \pm 0.32$ & $\textbf{99.65} \pm 0.23$ & $98.86 \pm 0.51$ & $99.52 \pm 0.19$\\  
\hline Waveform		& $75.58 \pm 0.58$ & $69.21 \pm 3.17$ & \cellcolor{orange!15}$96.16 \pm 0.58$ & $\textbf{97.10} \pm 0.53$ & $93.47 \pm 1.41$ & $95.81 \pm 0.57$\\ 
\hline Mushroom 	& $88.63 \pm 1.03$ & $98.47 \pm 0.38$ & \cellcolor{orange!15}$\textbf{99.99} \pm 0.00$ & $99.89 \pm 0.69$ & $99.08 \pm 2.29$ & $99.97 \pm 0.02$\\ 
\hline Pendigits 	& $95.34 \pm 0.45$ & $95.84 \pm 0.29$ & \cellcolor{orange!15}$99.66 \pm 0.11$ & $\textbf{99.81} \pm 0.10$ & $99.22 \pm1.78$ &$99.19 \pm 1.14$\\ 
\hline Adult 		& $73.33 \pm 0.65$ & $59.42 \pm 3.70$ & \cellcolor{orange!15}$\textbf{92.37} \pm 0.34$ & $77.32 \pm 10.93$ & $84.25 \pm 5.66$ & $92.32 \pm 0.34$ \\ \hline 
\hline Average		& $80.21$ & $81.96$ & \cellcolor{orange!25}$\textbf{94.44}$ & $93.81$ & $92.74$ & $94.29$\\\hline
\end{tabular}
\smallskip
\caption{Mean performance and standard deviation for the TEST set using a 10x10 folds cross-validation process} \label{AUC}
\end{center}
\vspace{-10mm}
\end{table}

\subsection{Results}
\label{results}
\subsubsection{Performance evaluation: }

Table \ref{AUC} presents the predictive performance of \texttt{LMT}, \texttt{NBT}, \texttt{SNB}, our algorithm $\texttt{PKM}_{\texttt{MV}}$, $\texttt{PKM}_{\texttt{SNB}}$ and the baseline  $\texttt{KM}_{\texttt{MV}}$ using the ACC (accuracy) and the AUC criteria (presented as a {\%}). These results show the very good prediction performance of the $\texttt{PKM}_{\texttt{SNB}}$ algorithm. Its performance is indeed comparable to those of \texttt{LMT} and \texttt{SNB} which are the strongest ones. In addition, the use of local classifiers (algorithm $\texttt{PKM}_{\texttt{SNB}}$) provides a clear advantage over the use of the majority vote in each cluster as done in $\texttt{PKM}_{\texttt{MV}}$. Surprisingly, $\texttt{PKM}_{\texttt{SNB}}$ exhibits slightly better results than \texttt{SNB} while both use naive Bayes classifiers locally and $\texttt{PKM}_{\texttt{SNB}}$ is hampered by the fact that $K = J$, the number of classes. Better performance are expected when $K \geq J$. Finally, $\texttt{PKM}_{\texttt{SNB}}$ appears to be slightly superior to \texttt{SNB}, particularly  for the datasets which contain highly correlated features, for instance the PenDigits database.

\vspace{-3mm}
\subsubsection{Discussion about local models, complexity and others factors: }
 
In Section II in the paragraph about label prediction in predictive clustering, we proposed a list of desirable properties for the local prediction models used in each cluster. We come back to these items denoted from (i) to (vi) in discussing Tables \ref{AUC} and \ref{tab:other}: 
\begin{itemize}
   \item[i)] The performance in prediction are good even for the dataset Glass which contains only 214 instances (90\% for training in the 10x10 cross validation (therefore 193 instances)).
   \item[ii)] The robustness (ratio between the performance in test and training) is given in Table \ref{tab:other} for the Accuracy (ACC) and the AUC. This ratio indicates that there is no significant overfitting. Moreover, by contrast to methods described in \cite{LMT,Kohavi96scalingup} (about \texttt{LMT} and \texttt{NBT}) our algorithm does not require any cross validation for setting parameters.
   \item[iii)] The only user parameter is the number of cluster (in this paper $K=J$). This point is crucial to help a non-expert to use the proposed method.
   \item[iv)] The preprocessing complexity (step 1 of Algorithm \ref{algo}) is $O(d \, m \log m)$, the k-means has the usual complexity $O(d \, m \, J \, t)$ and the complexity for the creation of the local models is $O(d \, m^* \log m*)+O(K(d \, m^* \log d m^*))$ where $d$ is the number of variables,  $m$ the number of instances in the training dataset, $m^*$ is the average number of instances belonging to a cluster. Therefore a fast training time is possible as indicated in Table \ref{tab:other}  with time given in seconds (for a PC with Windows 7 enterprise and a CPU : Intel Core I7 6820-HQ 2.70 GHz). 
   \item[v)] Only clusters where the information is sufficient to beat the majority vote contain local model.  Table \ref{tab:other} gives the percentage of pure clusters obtained at the end of the convergence the K-Means and the percentage of clusters with a local model (if not pure) when performing the 10x10 cross validation (so over 100 results).
   \item[vi)] Finally, the interpretation of the $\texttt{PKM}_{\texttt{SNB}}$ model is based on a two-level analysis. The first analysis consists in analyzing the profile of each cluster using histograms. A visualisation of  the average profile of the overall population (each bar representing the percentage of instances having a value of the corresponding interval) and  the average profile of a given cluster allows to understand why a given instance belongs to a cluster. Then locally to a cluster the variable importance of the local classifier (the weights, $W_f$, in the SNB classifier)  gives a local interpretation.
\end{itemize} 
\vspace{-5mm}

\begin{table}[!ht]
\fontsize{6}{9}\selectfont
\begin{center}
\begin{tabular}{|c|c|c|c|c|c|c||c||c|c|c|c|c|c|}
\hline  {\bf Datasets} & \multicolumn{2}{c|}{Robustness} & Training  & \multicolumn{3}{c|}{Local models} & {\bf Datasets} & \multicolumn{2}{c|}{Robustness} & Training  & \multicolumn{3}{c|}{Local models} \\
(\#intances)  			   & ACC & AUC & Time (s)& (1) & (2)  & (3) & (\#intances)  			   & ACC & AUC & Time (s)& (1) & (2)  & (3)  \\\hline
Glass & 0.98	& 0.99 & 0.07  & 40.83 & 23.17 & 36.0 & Waveform & 0.97	& 0.99 & 0.73  & 0.00 & 98.00 & 2.00\\\hline
Pima	& 0.95	& 0.94 & 0.05  & 00.00 & 88.50 & 11.5 & Mushroom 	& 1.00	& 1.00 & 0.53  & 50.00 & 50.00 & 0\\\hline
Vehicle& 0.95	& 0.97 & 0.14  & 07.50 & 92.25 & 0.25 & Pendigits 	& 0.99	& 1.00 & 1.81  & 0.00 & 100.00 & 0\\\hline
Segment.	& 0.98	& 1.00 & 0.85  & 28.28 & 66.28 & 5.44 & Adult 	 & 1.00	& 1.00 & 3.57   & 0.00 & 100.00 & 0\\\hline
\end{tabular}
\smallskip
\caption{Elements for discussion about local models. (1) Percentage of pure clusters; (2) Percentage of non-pure clusters with a local model; (3) Percentage of non-pure clusters without a local model.} \label{tab:other}
\end{center}
\vspace{-11mm}
\end{table}

The results of our experiments and the elements (i) to (vi) show that the algorithm $\texttt{PKM}_{\texttt{SNB}}$ is interesting with regards to several aspects. ($1$) Its predictive performance are comparable to those of the best competing supervised classification methods,  ($2$) it doesn't require cross validation, ($3$) it deals with the missing values, ($4$)  it operates a features selection both  in the  clustering step and during the building of the local models.  Finally, ($5$) it groups the categorical features into modalities, thus allowing one to avoid using a complete disjunctive coding which involves the creation of large vectors. Otherwise this disjunctive coding could complicate the interpretation of the obtained model.

The reader may find a supplementary material here: \url{https://bit.ly/2T4VhQw} or here: \url{https://bit.ly/3a7xmFF}. It gives a detailed example about the interpretation of the results and some comparisons to others predictive clustering algorithms as COBRA or MPCKmeans.


\vspace{-2mm}
\section{Conclusion and perspectives}
\label{conclusion}
\vspace{-2mm}

We have shown how to modify a distance-based clustering technique, such as k-means, into a predictive clustering algorithm. Moreover the learned representation could be used by other clustering algorithms. The resulting algorithm $\texttt{PKM}_{\texttt{SNB}}$ exhibits strong predictive performances most of the time as the state of the art but with the benefit of not having any parameters to adjust and therefore no cross validation to compute. The suggested algorithm is also a good support for interpretation of the data.  Better performances can still be expected when the number of clusters is higher than the number of classes. One goal of a work in progress it to find a method that would automatically discover the optimal number of clusters.  In addition, we are developing a tool to help visualize the results allowing the navigation between clusters in order to view easily the average profiles and the importance of the variables locally for each cluster.

\bibliographystyle{splncs04}
\bibliography{biblio_v2}

\begin{thebibliography}{10}
\providecommand{\url}[1]{\texttt{#1}}
\providecommand{\urlprefix}{URL }
\providecommand{\doi}[1]{https://doi.org/#1}

\bibitem{Al-Harbi2006}
Al-Harbi, S.H., Rayward-Smith, V.J.: Adapting k-means for supervised
  clustering. Applied Intelligence  \textbf{24}(3),  219--226 (2006)

\bibitem{Arthur2007}
Arthur, D., Vassilvitskii, S.: K-means++: The advantages of careful seeding.
  In: Proceedings of the Eighteenth Annual ACM-SIAM Symposium on Discrete
  Algorithms. pp. 1027--1035 (2007)

\bibitem{WHI}
Been~Kim, Kush R.~Varshney, A.W.: Workshop on human interpretability in machine
  learning (whi 2018). In: Proceedings of the 2018 ICML Workshop (2018)

\bibitem{Bilenko}
Bilenko, M., Basu, S., Mooney, R.J.: Integrating constraints and metric
  learning in semi-supervised clustering. In: Proceedings of the Twenty-first
  International Conference on Machine Learning (ICML) (2004)

\bibitem{bookpct}
Blockeel, H., Dzeroski, S., Struyf, J., Zenko, B.: Predictive Clustering.
  Springer-Verlag New York (2019)

\bibitem{Bouchard2004}
Bouchard, G., Triggs, B.: The tradeoff between generative and discriminative
  classifiers. In: IASC International Symposium on Computational Statistics
  (COMPSTAT). pp. 721--728 (2004)

\bibitem{BoulleJMLR05}
Boull\'e, M.: A {B}ayes optimal approach for partitioning the values of
  categorical attributes. Journal of {M}achine {L}earning {R}esearch
  \textbf{6},  1431--1452 (2005)

\bibitem{BoulleML06}
Boull\'e, M.: {MODL}: a {B}ayes optimal discretization method for continuous
  attributes. Machine {L}earning  \textbf{65}(1),  131--165 (2006)

\bibitem{BoulleJMLR07}
Boull\'e, M.: Compression-based averaging of selective naive {B}ayes
  classifiers. Journal of {M}achine {L}earning {R}esearch  \textbf{8},
  1659--1685 (2007)

\bibitem{Cevikalp}
Cevikalp, H., Larlus, D., Jurie, F.: A supervised clustering algorithm for the
  initialization of rbf neural network classifiers. In: Signal Processing and
  Communication Applications Conference (June 2007),
  \url{http://lear.inrialpes.fr/pubs/2007/CLJ07}

\bibitem{Eick04}
Eick, C.F., Zeidat, N., Zhao, Z.: Supervised clustering - algorithms and
  benefits. In: International Conference on Tools with Artificial Intelligence.
  pp. 774--776 (2004)

\bibitem{flach2012machine}
Flach, P.: Machine learning: the art and science of algorithms that make sense
  of data. Cambridge University Press (2012)

\bibitem{Hand2001}
Hand, D.J., Yu, K.: Idiot's bayes-not so stupid after all? International
  Statistical Review  \textbf{69}(3),  385--398 (2001)

\bibitem{Kohavi96scalingup}
Kohavi, R.: Scaling up the accuracy of naive-bayes classifiers: a decision-tree
  hybrid. In: International Conference on Data Mining. pp. 202--207. AAAI Press
  (1996)

\bibitem{LMT}
Landwehr, N., Hall, M., Frank, E.: Logistic model trees. Mach. Learn.
  \textbf{59}(1-2) (2005)

\bibitem{Langley1994}
Langley, P., Sage, S.: Induction of selective bayesian classifiers. In:
  Proceedings of the Tenth International Conference on Uncertainty in
  Artificial Intelligence. pp. 399--406. Morgan Kaufmann Publishers Inc., San
  Francisco, CA, USA (1994)

\bibitem{LemaireIJCNN2015initialisation}
Lemaire, V., Alaoui~Ismaili, O., Cornu\'ejols, A.: An initialization scheme for
  supervized k-means. In: International Joint Conference on Neural Networks
  (2015)

\bibitem{Lichman2013}
Lichman, M.: {UCI} machine learning repository (2013)

\bibitem{Manning2008}
Manning, C.D., Raghavan, P., Schütze, H.: Introduction to Information
  Retrieval. Cambridge University Press, New York (2008)

\bibitem{Meila1998}
Meil\u{a}, M., Heckerman, D.: An experimental comparison of several clustering
  and initialization methods. In: Conference on Uncertainty in Artificial
  Intelligence. pp. 386--395. Morgan Kaufmann Publishers Inc. (1998)

\bibitem{Milligan1988}
Milligan, G.W., Cooper, M.C.: {A study of standardization of variables in
  cluster analysis}. Journal of Classification  \textbf{5}(2),  181--204 (1988)

\bibitem{SalperwyckIJCNN2011learning}
Salperwyck, C., Lemaire, V.: Learning with few examples: An empirical study on
  leading classifiers. In: International Joint Conference on Neural Networks
  (2011)

\bibitem{cobra2}
Van~Craenendonck, T., Dumancic, S., Van~Wolputte, E., Blockeel, H.: {COBRAS:}
  fast, iterative, active clustering with pairwise constraints. In: Proceedings
  of Intelligent Data Analysis (2018)

\end{thebibliography}

\end{document}